# Real-time Indian Sign Language (ISL) Recognition

Kartik Shenoy, Tejas Dastane, Varun Rao, Devendra Vyavaharkar
Department of Computer Engineering,
K. J. Somaiya College of Engineering, University of Mumbai
Mumbai, India
kartik.s@somaiya.edu, tejas.dastane@somaiya.edu, varun.rao@somaiya.edu, devendra.v@somaiya.edu

*Abstract*—This paper presents a system which can recognise hand poses & gestures from the Indian Sign Language (ISL) in real-time using grid-based features. This system attempts to bridge the communication gap between the hearing and speech impaired and the rest of the society. The existing solutions either provide relatively low accuracy or do not work in real-time. This system provides good results on both the parameters. It can identify 33 hand poses and some gestures from the ISL. Sign Language is captured from a smartphone camera and its frames are transmitted to a remote server for processing. The use of any external hardware (such as gloves or the Microsoft Kinect sensor) is avoided, making it user-friendly. Techniques such as Face detection, Object stabilisation and Skin Colour Segmentation are used for hand detection and tracking. The image is further subjected to a Grid-based Feature Extraction technique which represents the hand's pose in the form of a Feature Vector. Hand poses are then classified using the k-Nearest Neighbours algorithm. On the other hand, for gesture classification, the motion and intermediate hand poses observation sequences are fed to Hidden Markov Model chains corresponding to the 12 pre-selected gestures defined in ISL. Using this methodology, the system is able to achieve an accuracy of 99.7% for static hand poses, and an accuracy of 97.23% for gesture recognition.

## I. INTRODUCTION

Indian Sign Language (ISL) is a sign language used by hearing and speech impaired people to communicate with other people. The research presented in this paper pertains to ISL as defined in the Talking Hands website [1]. ISL uses gestures for representing complex words and sentences. It contains 33 hand poses including 10 digits, and 23 letters. Amongst the letters in ISL, the letters 'h', 'j' are represented by gestures and the letter 'v' is similar to digit 2. The system is trained with the hand poses in ISL as shown in Fig. 1.

Most people find it difficult to comprehend ISL gestures. This has created a communication gap between people who understand ISL and those who do not. One cannot always find an interpreter to translate these gestures when needed. To facilitate this communication, a potential solution was implemented which would translate hand poses and gestures from ISL in real-time. It comprises of an Android smartphone camera to capture hand poses and gestures, and a server to process the frames received from the smartphone camera. The purpose of the system is to implement a fast and accurate recognition technique.

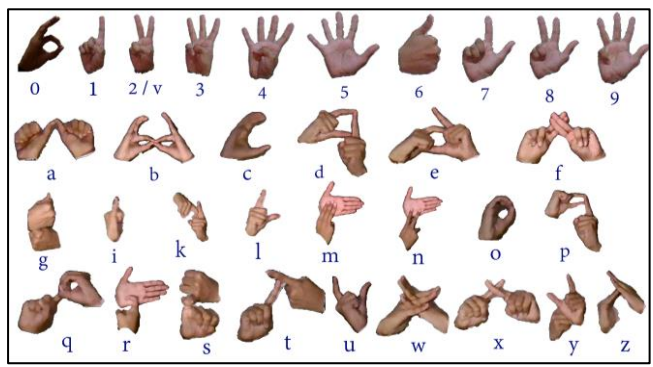

Fig. 1. Hand poses in ISL

The system described in this paper successfully classifies all the 33 hand poses in ISL. For the initial research, gestures containing only one hand was considered. The solution described can be easily extended to two-handed gestures. In the next section of this paper, the related work pertaining to sign language translation is discussed. Section III explains the techniques used to process each frame and translate a hand pose/gesture. Section IV discusses the experimental results after implementing the techniques discussed in section III. Section V describes the Android application developed for the system that enables real-time Sign Language Translation. Section VI discusses the future work that can be carried out in ISL translation.

## II. RELATED WORK

There has been considerable work in the field of Sign Language recognition with novel approaches towards gesture recognition. Different methods such as use of gloves or Microsoft Kinect sensor for tracking hand, etc. have been employed earlier. A study of many different existing systems has been done to design a system that is efficient and robust than the rest.

A Microsoft Kinect sensor is used in [2] for recognising sign languages. The sensor creates depth frames; a gesture is viewed as a sequence of these depth frames. T. Pryor *et al* [3] designed a pair of gloves, called SignAloud which uses embedded sensors in gloves to track the position and movement of hands, thus converting gestures to speech. R. Hait-Campbell *et al* [4] developed MotionSavvy, a technology that uses Windows tablet and Leap Motion accelerator AXLR8R to recognise the hand, arm skeleton. Sceptre [5] uses Myo gesture-control armbands that provide accelerometer, gyroscope and electromyography (EMG) data for signs & gestures classification. These hardware solutions provide good accuracy but are usually expensive and are not portable. Our system eliminates the need of external sensors by relying on an Android phone camera.

Now for software-based solutions, there are coloured glove based [6, 7] and skin colour-based solutions. R. Y. Wang et al [6] have used multi-coloured glove for accurate hand pose reconstruction but the sign demonstrator, while demonstrating the sign language, has to wear this each time. Skin colour-based solutions may use RGB colour space with some motion cues [8] or HSV [9, 10, 11], YCrCb [12] colour space for luminosity invariance. G. Awad et al [13] have used the initial frames of the video sequence to train the SVM for skin colour variations for the further frames. But to speed up the skin segmentation, they have used Kalman filter for further prediction of position of skin coloured objects thus reducing the search space. Z. H. Al-Tairi et al [14] have used YUV and RGB colour space for skin segmentation and the colour ranges that they have used handles good variation of people's races.

After obtaining segmented hand image, A. B. Jmaa et al [12] have used the rules defined in the hand anthropometry study of comparative measurements of human body for localizing and eliminating the palm. They have then used the rest of the segmented image containing only fingers to create skin-pixel histogram with respect to palm centroid. This histogram is fed to decision tree classifier. In [15], from the segmented hand image, hand contour was obtained, which was then used for fitting a convex hull and convexity defects were found out. Using this, the fingers were identified and the angles between the adjacent ones were determined. This feature set of angles was fed to SVM for classification. [10] have used distance transform to identify hand centroid followed by elimination of palm and using angles between fingers for classification. Fourier Descriptors have been used to describe hand contours by [8, 16]. [16] has used RBF on these Fourier Descriptors for hand pose classification. S. C. Agarwal et al [17] have used a combination of geometric features (eccentricity, aspect ratio, orientation, solidity), Histogram of Oriented Gradients (HOG) and Scale Invariant Fourier Transform (SIFT) key points as feature vectors. The accuracy obtained using geometric features goes really low when number of hand poses increases. [18] has used Local Binary Patterns (LBP) as features. Our paper is mainly inspired from [9]. They have trained the k-NN model using the binary segmented hand images directly. This technique provides great speed when combined with fast indexing methods, thus making it suitable for real-time applications. But to handle the variations in hand poses, more data needs to be captured. With the use of grid-based features in our system, the model will become more user-invariant.

For gesture recognition, hand centroid tracking is done which provides motion information [16]. Gesture recognition can be done using the Finite State Machine [19] which has to be defined for each gesture. C. Y. Kao et al [20] have used 2 hand gestures for training HMM that will be used for gesture recognition. They defined directive gestures such as up, left, right, down for their 2 hands and a time series of these pairs was input to the HMM for gesture recognition. C. W. Ng et al [16] used a combination of HMM and RNN classifiers. The HMM Gesture recognition that we have used in our system is mainly inspired from [16]. They were using 5 hand poses and the same 4 directive gestures. This 9-element vector was used as input to the HMM classifier. Training of HMM was done using Baum-Welch re-estimation formulas.

III. IMPLEMENTATION

Using an Android smartphone, gestures and signs performed by the person using ISL are captured and their frames are transmitted to the server for processing. To make the frames ready for recognition of gestures and hand poses, they need to be pre-processed. The pre-processing first involves face removal, stabilisation and skin colour segmentation to remove background details and later morphology operations to reduce noise. The hand of the person is extracted and tracked in each frame. For recognition of hand poses, features are extracted from the hand and fed into a classifier. The recognised hand pose class is sent back to the Android device. For classification of hand gestures, the intermediate hand poses are recognised and using these recognised poses and their intermediate motion, a pattern is defined which is represented in tuples. This is encoded for HMM and fed to it. The gesture whose HMM chain gives the highest score with forward-backward algorithm is determined to be the recognized gesture for this pattern. An overview of this process is described in Fig. 2.

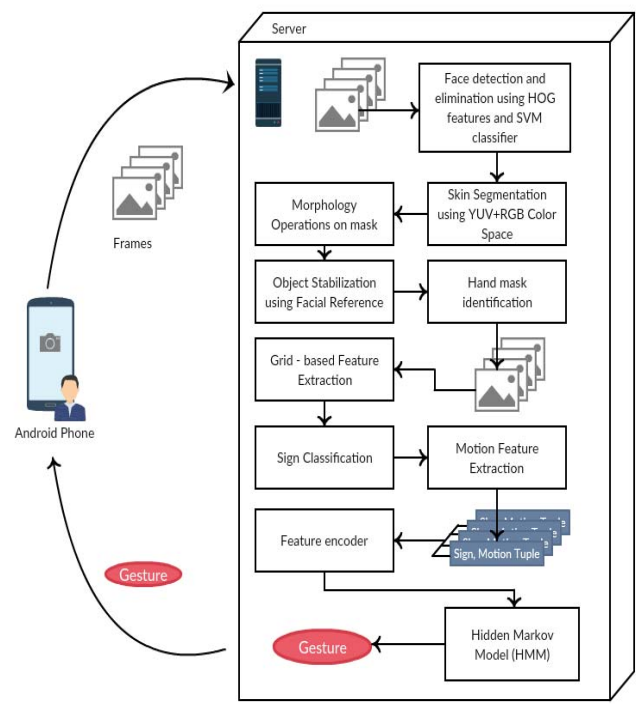

Fig. 2. Flow diagram for Gesture Recognition.

A. *Dataset used*

For the digits 0 to 9 in ISL, an average of 1450 images per digit were captured. For letters in ISL excluding 'h', 'j' and 'v', about 300 images per letter were captured. For the 9 gesture-related intermediate hand poses such as Thumbs_Up, Sun_Up, about 500 images per pose were captured. The dataset contains a total of 24,624 images. All the images consist of the sign demonstrator wearing a full sleeve shirt.

Most of these images were captured from an ordinary webcam and a few of them were captured from a smartphone camera. The images are of varying resolutions. For training HMMs, 15 gesture videos were captured for each of the 12 one-handed pre-selected gestures defined in [1] (After, All The Best, Apple, Good Afternoon, Good Morning, Good Night, I Am Sorry, Leader, Please Give Me Your Pen, Strike, That is Good, Towards). These videos have slight variations in sequences of hand poses and hand motion so as to make the HMMs robust. These videos were captured from a smartphone camera and also involve the sign demonstrator wearing a full sleeve shirt.

### B. Pre-processing

#### 1) Face detection and elimination

The hand poses and gestures in ISL can be represented by particular movement of hands, and facial features are not necessary. Also, face of the person creates an issue during hand extraction process. To resolve this issue, face detection was carried out using Histogram of Oriented Gradients (HOG) descriptors followed by a linear SVM classifier. It uses an image pyramid and sliding window to detect faces in an image, as described in [21]. HOG feature extraction combined with a linear classifier reduces false positive rates by more than an order of magnitude than the best Haar wavelet-based detector [21]. After detection of face, the face contour region is identified, and the entire face-neck region is blackened out, as demonstrated in Fig. 3.

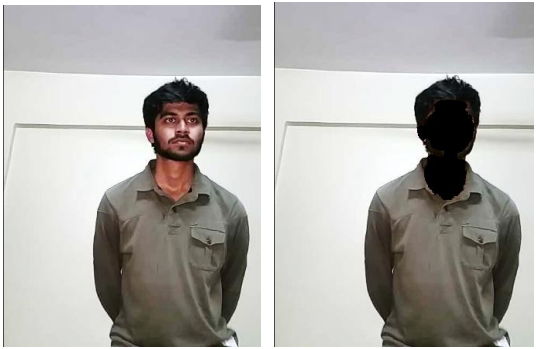

Fig. 3. Face detection and elimination operation

#### 2) Skin colour segmentation

To identify skin-like regions in the image, the YUV and RGB based skin colour segmentation is used, which provides great results. This model has been chosen since it gives the best results among the options considered: HSV, YCbCr, RGB, YIQ, YUV and few pairs of these colour spaces [14]. The frame is converted from RGB to YUV colour space using the equation mentioned in [22]. This is specified in equation (1).

$$\begin{pmatrix} Y \\ U \\ V \end{pmatrix} = \begin{pmatrix} +0.299 & +0.587 & +0.114 \\ -0.147 & -0.289 & +0.436 \\ +0.615 & -0.515 & -0.100 \end{pmatrix} \cdot \begin{pmatrix} R \\ G \\ B \end{pmatrix} + \begin{pmatrix} 0 \\ 128 \\ 128 \end{pmatrix} \quad (1)$$

[14] mentions the criteria for determining if pixel is to be classified as skin using the following conditions:

$$\begin{cases} 80 < U < 130 \\ 136 < V < 200 \\ V > U \\ R > 80 \ \& \ G > 30 \ \& \ B > 15 \\ |R - G| > 15 \end{cases} \quad (2)$$

The segmentation mask thus obtained contains less noise and less false positive results. An illustration of the segmentation mask is shown in Fig. 4.

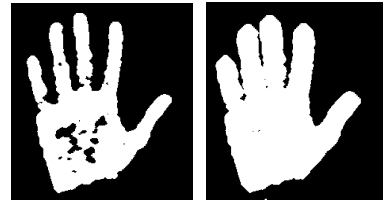

Fig. 4. Skin segmentation mask and effect of morphology operations. (Left) Segmentation mask; (Right) Mask after application of Morphology operations.

#### 3) Morphology operations

Morphology operations were performed to remove any noise generated after skin colour segmentation. There are 2 types of errors in skin colour segmentation:

1. Non-skin pixels classified as skin
2. Skin pixels classified as non-skin

Morphology involves 2 basic sub-operations:

1. Erosion: Here, the active areas in the mask (which are white) are reduced in size
2. Dilation: Here, the active areas in the mask (which are white) are increased in size

Morphology Open operation handles the $1^{st}$ error. It involves erosion followed by dilation. The $2^{nd}$ error is handled by Morphology Close operation. This involves dilation followed by erosion. The result of applying morphology operations can be seen in Fig. 4.

#### 4) Object stabilisation using Facial reference.

For tracking hand motion accurately, a steady camera position is desired. Movement of the camera caused by shaky hands is common. If the sign demonstrator, that is, the person using ISL does not move his hand but the person capturing the video shakes his hand, false movements will get detected. This problem is tackled using object stabilisation.

Under the assumption that a person's face is always included in the gesture video, the face of the sign demonstrator is tracked to stabilise hand movements. The tracker is initialised with the co-ordinates which were extracted from Face Detection before removing the face. The tracker detects the location of the facial blob and shifts the entire frame in the opposite direction of the detected motion of facial coordinates. The system uses the Kernelized Correlation Filter (KCF) tracker implemented in the OpenCV library to track the face in each frame. The operation of tracking is performed on the image before the face is blackened. Fig. 5 demonstrates object stabilisation.

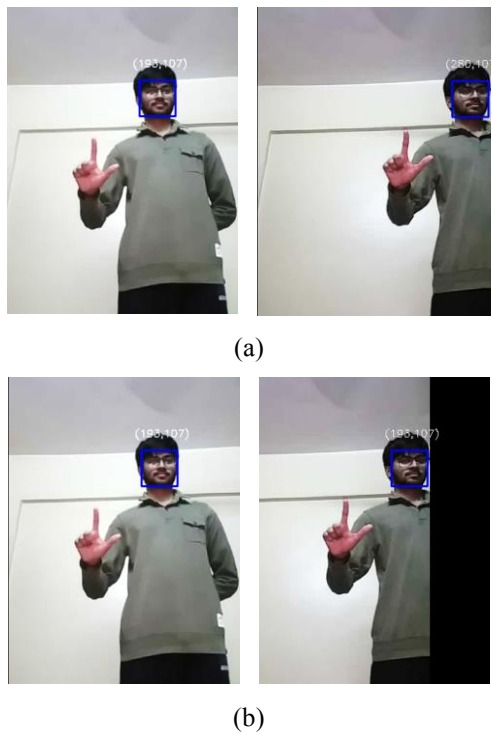

Fig. 5. (a) Actual motion of camera. The subject's position with respect to the frame changes. (b) Effect after stabilisation – The position of subject with respect to the frame remains constant.

*C. Hand extraction and tracking*

As all ISL hand poses and gestures can be represented using hand movements, hand extraction and tracking is important part of the system. After pre-processing each frame, a black and white image is obtained, where white areas represent skin. These skin areas do not contain facial regions. They contain parts of the hand and other skin-like parts in the original image. Since each frame contains only 1 hand (the other hand is not visible) or both hands are touching each other, the only prominent contour present in the frame will be the person's hand. Thus, areas of all contours in the frame are calculated and the contour with the largest area is extracted. Since the only prominent contour left is hand contour, the extracted contour is the hand region. Fig. 6 illustrates the importance of face elimination operation. If face was not eliminated, the face region would have been the largest contour area in the image, thus, being classified as hand as shown in Fig. 6.

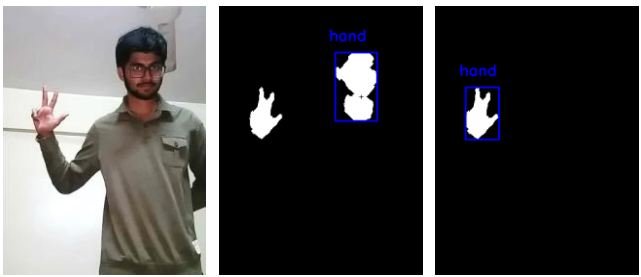

Fig. 6. Importance of eliminating face before hand extraction.

For tracking hand motion, centroid of hand is calculated in each frame. If there is movement of hand, the co-ordinates of centroid of hand will change. Slope of the line formed by the centroid of hand in the current frame and centroid of hand in previous frame is then determined. Depending upon the value of slope, the motion of hand was determined as follows:

- If -1 < slope < 1 and difference between x co-ordinates of both centroids is positive, the hand moved leftwards.
- If -1 < slope < 1 and difference between x co-ordinates of both centroids is negative, the hand moved rightwards.
- If |slope| > 1 and difference between y co-ordinates of both centroids is positive, hand moved upwards.
- If |slope| > 1 and difference between y co-ordinates of both centroids is negative, hand moved downwards.

The above slope-based motion determination is illustrated in Fig. 7. It is to be noted that the hand motion observed by the camera will be opposite to the actual motion performed by the sign demonstrator. The system uses the OpenCV library to calculate area of contour and centroid of hand contour.

To reduce noise during hand tracking, an imaginary circle with a 20-pixel radius around the previous hand centroid pixel is considered. If the new centroid lies within this radius, this shift is termed as noise and motion is neglected. The previous co-ordinate considered during comparison is not updated in this case. If the current centroid lies outside the 20-pixel threshold, this shift is considered as movement of hand. In the subsequent frames, the radius is set to 7 pixels instead of 20 pixels until there is no movement after which the radius is restored to 20 pixels. This use of an imaginary circle reduces noise to a greater extent and gives highly accurate tracking of hand movements.

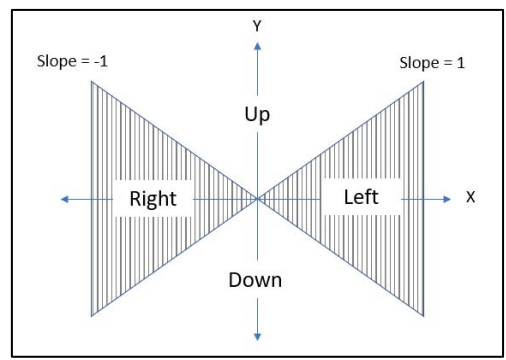

Fig. 7. Determining hand motion using slope.

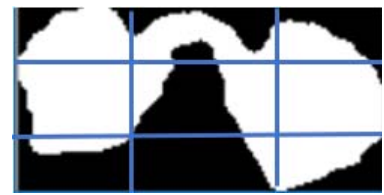

Fig. 8. The hand pose 'a' in ISL fragmented by a 3x3 grid.

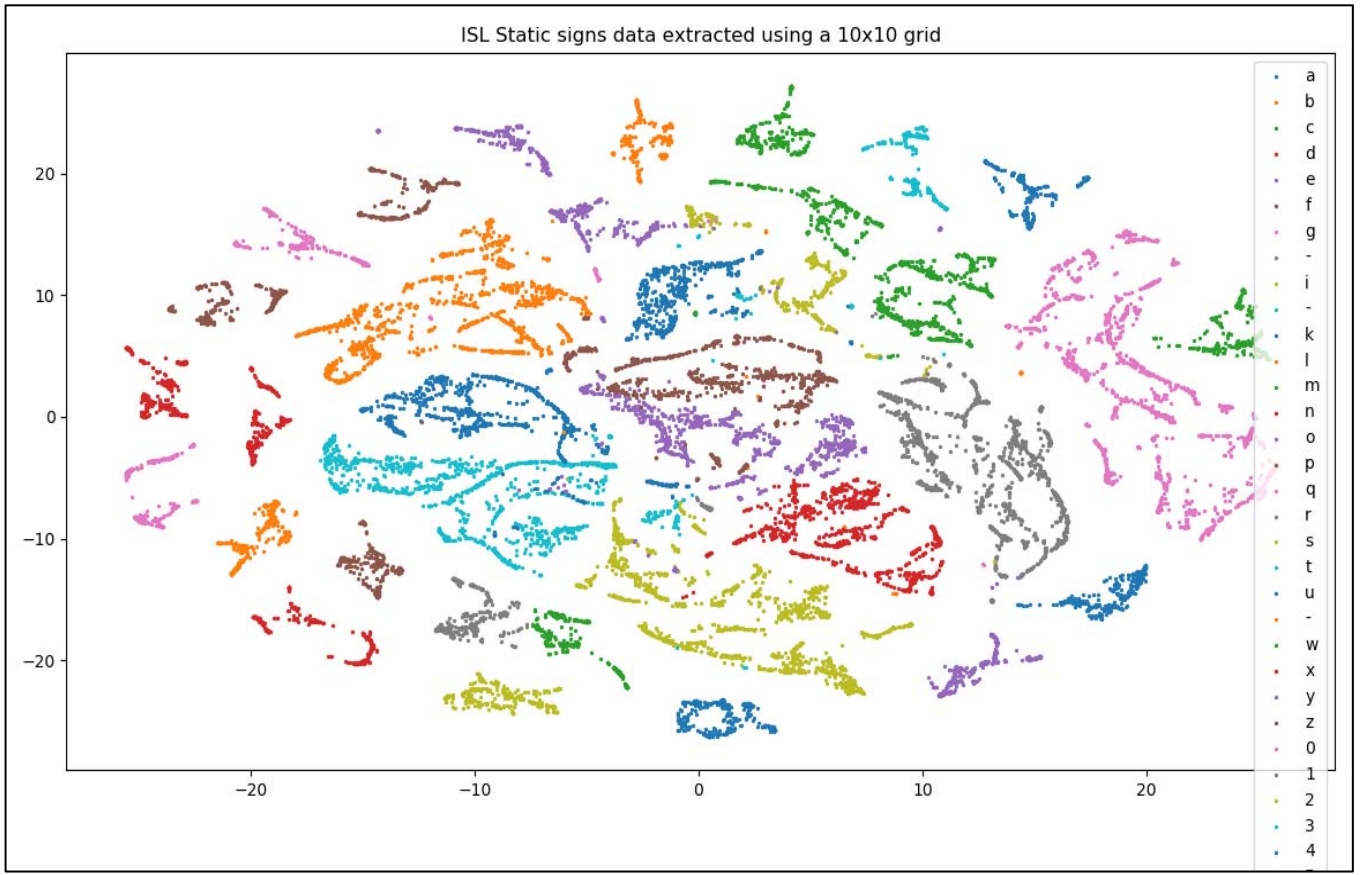

Fig. 9. ISL hand poses' data visualised using PCA and TSNE dimensionality reduction.

### D. Feature extraction using Grid-based fragmentation technique

Using an M x N grid, extracted hand sample is fragmented into M*N blocks. Using this grid, a feature vector is obtained containing M*N feature values where each block provides a feature value. In each block, the feature value is calculated as the percentage of hand contour present. This is specified in equation (3).

$$Feature\ value = \frac{Area\ of\ hand\ contour}{Area\ of\ fragment} \quad (3)$$

If no contour is found, the feature value is 0. In Fig. 8, a 3 x 3 grid is constructed on a sample for the purpose of illustration. The advantage of using this approach is the features generated vary with the orientation of each hand pose. Different hand poses occupy different number of grids and different fragment areas. The feature vector thus accurately represents the shape and position of the hand. Using these M*N features, each hand pose is represented by different clusters.

Fig. 9 supports these arguments. The data visualised in Fig. 9 was subjected to a 10 x 10 grid and 100 features were extracted per sample. Using Principle Component Analysis (PCA) and t-Distributed Stochastic Neighbour Embedding (t-SNE) [23], the dimensionality of this data was reduced from 100 to 2. This was done by first applying PCA to reduce the dimensionality to 40 features and then t-SNE to reduce it further to 2. As seen in Fig. 9, separate clusters representing different orientations of each hand pose can be seen after visualising the extracted features.

### E. Classification

*1) Recognition of ISL Hand poses using k-NN*

After observing the graph plotted in Fig. 9, it can be observed that data is organised into clusters. There is more than one cluster for the same hand pose. For classification, an algorithm was needed which can distinguish clustered data efficiently. K-Nearest Neighbours (k-NN) was found suitable for such distribution of data. Hand extracted from each frame of the live feed is subject to feature extraction using the previously discussed Grid-based fragmentation. This sample is then represented in an M*N dimensional feature map.

Using Euclidean distance as a distance metric, k nearest samples, which are fitted previously in the classifier, are computed. The distance computation can be performed using a brute force approach, wherein Euclidean distance between the sample and each fitted sample in the classifier is calculated and the k lowest distances are selected. Other optimal approaches include the KD-tree and Ball Tree. The most suitable approach for distance computation is dependent on size of the data. Brute force works well on small data size [24]. For low dimensional data, KD Tree works well whereas for high dimensional data, Ball Tree works best [24]. From these k nearest neighbours, the classifier selects the class occurring most frequently.

*2) Gesture Classification using HMM*

There are always some variations present while demonstrating gestures even if performed by the same sign demonstrator. For handling such variations, some kind of statistical model is needed. Hidden Markov Models (HMMs) are one type of statistical model that can handle such variations well [25]. There are two types of HMMs: continuous and discrete HMM. In continuous HMM, the number of observation symbols possible in each element of the observation sequences is infinite, but in discrete HMM, it is finite. The HMM can either be ergodic or left-to-right. In left-to-right HMM, the transition can occur only in 1 direction i.e. if the HMM moves to the next state it cannot go back to the previous state as shown in Fig. 10. But in ergodic HMM, transition is possible from any one state to any other state. The initial state probabilities ($\pi$) and the transition probabilities for left-to-right HMM have been shown in Fig. 10.

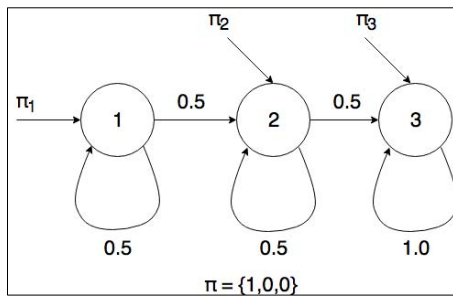

Fig. 10. HMM chain for gesture with 3 hidden states (E.g. Good Afternoon)

A human brain perceives gestures as a combination of a few intermediate hand poses and hand movements executed in a particular order. Using this idea, ISL Gestures are comprised of intermediate stationery hand poses and the hand movements between these. Thus, in this system, discrete left-to-right HMM has been used. This uses the segmented hand centroid motion and pose classification results for gesture classification of provided observation sequence as belonging to one of the 12 predefined gestures.

The input to this HMM is an observation sequence extracted from the video feed. The number of observation symbols possible is the sum of the number of directions tracked and the number of intermediate stationery hand poses with which the system is trained. In this system, there are 4 directions being tracked (as described in section III C) and the system is trained with 9 intermediate stationery hand poses such as 'Thumbs_Up', 'Sun_Up', 'Fist' as shown in Fig. 11.

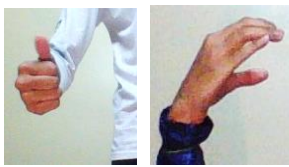

Fig. 11. Thumbs_Up and Sun_Up stationery hand poses

The intermediate hand pose recognition also uses Grid-based feature extraction. The intermediate hand pose recognition is similar to recognition of hand poses for recognizing letters and digits but is carried out only when there is no movement. Thus, the total number of observation symbols are 13 i.e. the observation sequence can have elements with values 0-12. At each frame, a tuple is generated of the form: <S, M> where M represents the Motion of the hand relative to the previous frame and S represents the pose classified if there was no motion detected. If motion is detected, the motion is mapped to the corresponding observation symbol ('Upwards':0, 'Rightwards':1, 'Leftwards':2, 'Downwards':3). If no motion is detected, the hand pose is accordingly mapped. This mapping is pre-decided. Therefore, for each frame there will be an observation symbol and the video sequence will generate an observation sequence which encodes motion and hand pose information. For example, the time series – [ <Thumbs_Up, None>, <Thumbs_Up, None>, <Thumbs_Up, None>, <None, Up>, <None, Up>, <None, Up>, <None, Up>, <Sun_Up, None>, <Sun_Up, None>, <Sun_Up, None> ] represents the gesture "Good Afternoon". After encoding this into an observation sequence, we get [ 4, 4, 4, 0, 0, 0, 5, 5, 5 ]. A few frames of this gesture are illustrated in Fig. 12.

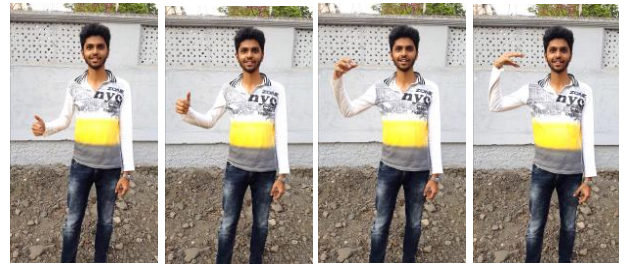

Fig. 12. 4 frames of Gesture "Good Afternoon"

The gesture recognition used in this system involves 12 HMM chains (one chain per gesture). The number of hidden states in each of these chains is determined by breaking the gesture into a combination of hand poses and the motion between them. For example, Good Afternoon gesture as shown in Fig. 12, can be said to have 3 states: 'Thumbs_Up' hand pose, 'Upwards' motion, 'Sun_Up' hand pose. All these HMM chains, being left-to-right HMMs, the initial state probabilities and the initial state transition probabilities are similar to the one shown in Fig. 10. For n hidden states in HMM, the state transition probability will be of order n x n and the emission probability matrix will be of order n x 13. The emission probability matrices were initialized with probabilities determined empirically by subjectively looking at the similarity between hand poses and the closest motions possible. This is to increase the chances of the HMM model converging into the global maxima after training.

After all the parameters as described here [26] are initialized as explained above, the estimation probabilities and state transition probabilities of HMM chains are trained using Baum-Welch algorithm [26, 27]. The HMM chains are trained by using a gesture database that was created, the details of which have been specified in section III A. After training the HMM chains, the new observation sequence is fed to these HMM chains, and the HMM chain giving the highest score with forward-backward algorithm [26] is determined to be the recognized gesture.

*3) Temporal Segmentation*

The gesture recognition module needs to be given video segments that correspond to the gesture only. Without temporal segmentation, continuous gesture recognition is not possible. We have used a rule where in if the hand goes out of frame, it would mark the end of the current gesture and gesture recognition will be performed over the currently obtained frame sequence from gesture. When the hand again comes within frame, it would mark the beginning of the new gesture. Temporal segmentation is achieved using this rule.

## IV. Experimental Results

The results discussed in this section were obtained from a personal computer with 8 GB of RAM, an Intel i7 processor with 4 virtual CPUs and a 2 GB NVIDIA GPU. The operating system used was Ubuntu 16.04.

For better classification of hand poses, an appropriate grid size had to be determined for fragmentation of hand and extraction of features. We applied 6 different grid sizes – 5x5, 10x10, 10x15, 15x15, 15x20 and 20x20 to extract features from the same training data. These features were then fitted into a k-NN classifier. Features were then extracted from the testing data using the grid sizes in consideration. The features extracted from the testing data were then classified using the k-NN classifier trained previously. An ideal grid size would form distinct clusters of hand poses so that the k-NN classifier can recognise them with high precision. To determine this ideal grid size, the accuracies of the trained k-NN classifiers were calculated using the testing data and were compared. The accuracy was calculated as specified in equation (4):

$$Accuracy = \frac{Number\ of\ samples\ classified\ correctly}{Total\ number\ of\ samples} \quad (4)$$

The comparison was plotted on a graph as shown in Fig. 13. From the graph, it can be inferred that the grid size of 10x10 gives the highest accuracy of 99.714%. In our implementation, the 10x10 grid was thus selected for extracting features from hand image. The average time required to extract features from an image of size 300x300 using a 10x10 grid was found to be about 1 milli-second. The testing data was 30% of the original dataset of hand poses.

Fig. 15 shows the confusion matrix of the k-NN classifier when tested with hand poses' data. It can be inferred from the confusion matrix that the model is able to distinguish between each hand pose accurately. The time taken for each frame has been tabulated in TABLE I. Thus, our application achieves a frame rate of 5.3 fps. After the frames have been processed and the time series has been extracted from the gesture frame sequence, the average time taken by HMM model (consisting of 12 HMM chains) for gesture classification is 3.7 ms.

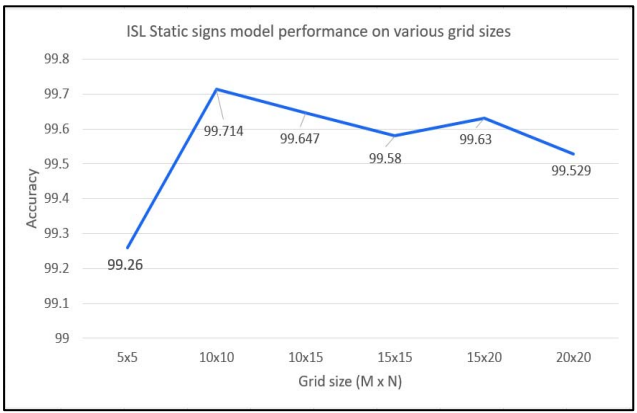

Fig. 13. Comparison of accuracy of k-NN classification on features extracted using various grid sizes on hand poses' data.

TABLE I. COMPUTATIONAL TIME FOR EACH PHASE

| Sr. No. | Phase | Time taken (in ms) |
|---|---|---|
| 1 | Data Transfer over WLAN | 46.2 |
| 2 | Skin Colour Segmentation and Morphological Operations | 12.3 |
| 3 | Face Detection and Elimination | 99.9 |
| 4 | Object Stabilization | 14.8 |
| 5 | Feature Extraction | 11.2 |
| 6 | Hand Pose Classification | 1.7 |
| | Average Time per Frame | 186.1 |

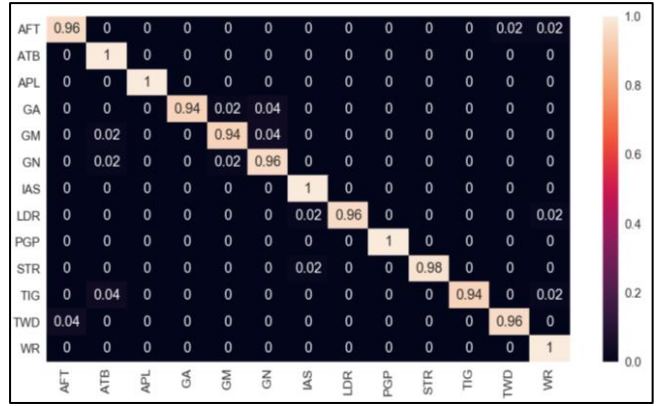

Fig. 14. Heatmap of Confusion matrix of ISL gestures.

12 gestures and their abbreviations that were considered for testing are as follows:

- After (AFT)
- All the Best (ATB)
- Apple (APL)
- Good Afternoon (GA)
- Good Morning (GM)
- Good Night (GN)
- I Am Sorry (IAS)
- Leader (LDR)
- Please Give Me Your Pen (PGP)
- Strike (STR)
- That is good (TIG)
- Towards (TWD)

Also, testing was done for Wrong Gestures (WR), that is, gestures that are not amongst the above listed. A 10x10 grid and a k-NN classifier was used to recognise the intermediate hand poses, similar to hand pose recognition. 50 real-time trials were performed for each gesture class in good lighting conditions with the sign demonstrator wearing a full sleeve shirt. Fig. 14 depicts a confusion matrix generated after obtaining the results of these trials. Each cell in the confusion matrix contains the percentage of trials that gave the corresponding output. It clearly shows that correct classification was obtained more than 94% of the time, and 97.2% on an average. It can thus be inferred from the results that the hand motion tracking and hand pose classification was accurate enough to generate appropriate time series for the Hidden Markov Models.

IEEE - 43488

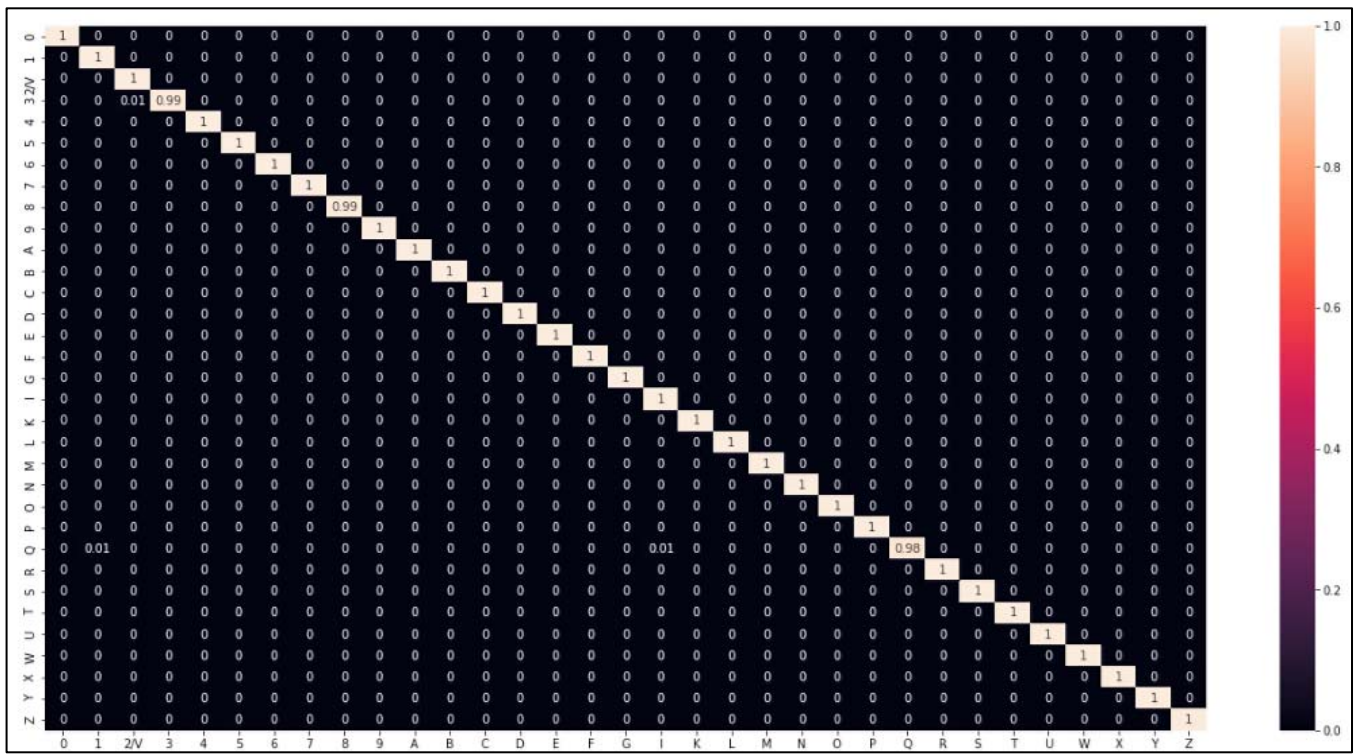

Fig. 15. Heatmap of Confusion matrix for k-NN classifier tested on ISL hand poses.

## V. THE APPLICATION

The system is implemented as an Android application. The application uses the smartphone's camera to capture the sign language used by the person. The frames were captured at a rate of 5 frames per second. Each frame is continuously sent to a remote server. The processing is performed at the server-side. After each pose or gesture is classified, the result is sent back to the application which is displayed in the top-portion. Currently, sockets are used to simulate a client-server connection. Fig. 16 shows screenshots of the application.

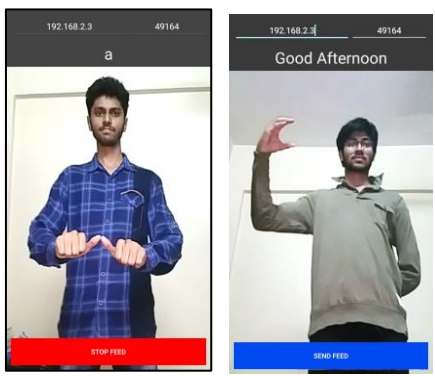

Fig. 16. (Left) Hand Pose 'a' recognized by the application; (Right) Gesture 'Good Afternoon' recognised by the application.

## VI. FUTURE WORK

Currently, only single-handed gestures in ISL were considered for research. With the use of advanced hand extraction algorithms, this approach can be extended to two-handed gestures as well. Also, with the use of Natural Language Processing algorithms, this system can be extended to recognise sentences in ISL, by recognition of multiple gestures in the same video capture.

Hand extraction is currently dependent on skin colour segmentation. This means that the hand extraction requires that the subject must wear full sleeve shirt for accurate recognition. Although the system could help the hearing and speech impaired community where full-sleeve shirts are frequently used, the system may not work in general conditions. This approach could be further extended using Object Detection techniques to extract hand region from the image. The only limitation in implementation of Object Detection techniques is requirement of a very wide variety of annotated hand samples so that it could detect hands in almost any position, orientation and background. The current approach also requires that the lighting conditions should be optimal – neither too dark nor too bright. The use of even better skin colour segmentation techniques which can perform well under a wider variety of lighting conditions can lead to better segmentation results and in turn aid in feature extraction.

## CONCLUSION

From the results, it can be inferred that the system presented in this paper is accurately able to track hand movements of the sign demonstrator using techniques such as Object Stabilisation, Face elimination, Skin colour Extraction and then Hand extraction. It can classify all 33 hand poses in ISL with an accuracy of 99.7%. The system was also able to classify 12 gestures with an average accuracy of 97.23%. The approach uses an HMM chain for each gesture and a k-NN model to classify each hand pose. The time required for recognition of hand pose is about 0.2s and that for gesture is 0.0037s. From the results, it can be concluded that the system can recognise hand poses and gestures in ISL with precision and in real-time. The system provides higher accuracy and faster recognition in sign language recognition than other approaches discussed in the literature. The approach discussed is inspired from various systems described in the Related Works section and utilises the pros discussed in their system to make the system more precise while classification. This approach is generic and can be extended to other single-handed and two-handed gestures. The system presented in this paper can also be extended to other Sign Languages, if dataset satisfying the current requirements of the system is available.